\documentclass[10pt,twocolumn,letterpaper]{article}
\usepackage{amsfonts}
\usepackage{iccv}
\usepackage{times}
\usepackage{epsfig}
\usepackage{graphicx}
\usepackage{amsmath}
\usepackage{amssymb}
\usepackage{subfigure}
\usepackage[pagebackref=true,breaklinks=true,letterpaper=true,colorlinks,bookmarks=false]{hyperref}

\iccvfinalcopy
\def\iccvPaperID{6107}

\ificcvfinal\fi
\input{tcilatex}
\begin{document}

\title{Multi-scale Processing of Noisy Images using Edge Preservation Losses}
\author{Nati Ofir\\
	BIU\\
	Ramat Gan, Israel\\
	{\tt\small natiofir@gmail.com}
	\and
	Yosi Keller\\
	BIU\\
	Ramat Gan, Israel\\
	{\tt\small yosi.keller@gmail.com }
}
\maketitle

\begin{abstract}
Noisy images processing is a fundamental task of computer vision. The first
example is the detection of faint edges in noisy images, a challenging
problem studied in the last decades. A recent study introduced a fast method
to detect faint edges in the highest accuracy among all the existing
approaches. Their complexity is nearly linear in the image's pixels and
their runtime is seconds for a noisy image. Their approach utilizes a
multi-scale binary partitioning of the image. By utilizing the multi-scale
U-net architecture, we show in this paper that their method can be
dramatically improved in both aspects of run time and accuracy. By training
the network on a dataset of binary images, we developed an approach for
faint edge detection that works in a linear complexity. Our runtime of a
noisy image is milliseconds on a GPU. Even though our method is orders of
magnitude faster, we still achieve higher accuracy of detection under many
challenging scenarios. In addition, we show that our approach to performing
multi-scale preprocessing of noisy images using U-net improves the ability
to perform other vision tasks under the presence of noise. We prove it on
the problems of noisy objects classification and classical image denoising.
We show that multi-scale denoising can be carried out by a novel edge
preservation loss. As our experiments show, we achieve high-quality results
in the three aspects of faint edge detection, noisy image classification and
natural image denoising.
\end{abstract}

\section{Introduction}

\begin{figure}[tbh]
	\centering
	\subfigure[]{\includegraphics[width=0.32\linewidth]{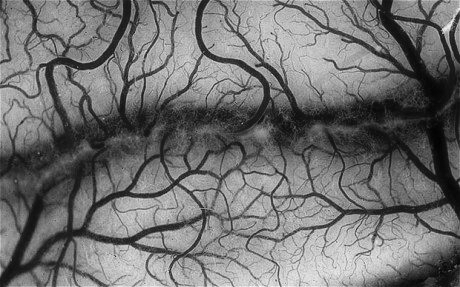}} %
	\subfigure[]{\includegraphics[width=0.32\linewidth]{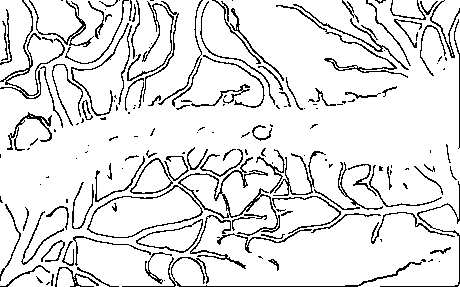}} %
	\subfigure[]{\includegraphics[width=0.32\linewidth]{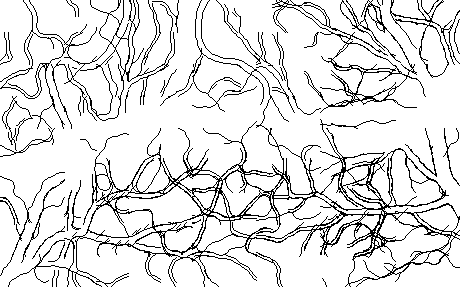}}
	\caption{Example of a medical image with many curved edges. (a) The original
		image. (b) The proposed DeepFaster approach results. (c) FastEdges
		\protect\cite{ofir2016fast} results. Both methods achieve high quality of
		detection while ours run in milliseconds and FastEdges runtime is more than
		seconds.}
	\label{fig:1}
\end{figure}
\begin{figure}[tbh]
	\centering
	\subfigure[]{\includegraphics[width=0.32\linewidth]{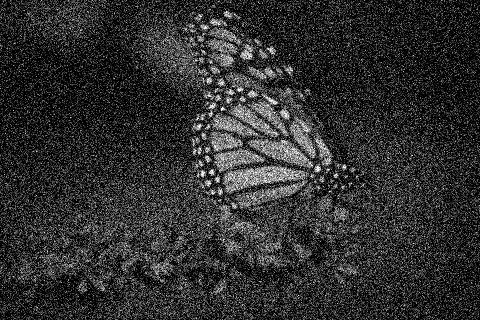}} %
	\subfigure[]{\includegraphics[width=0.32\linewidth]{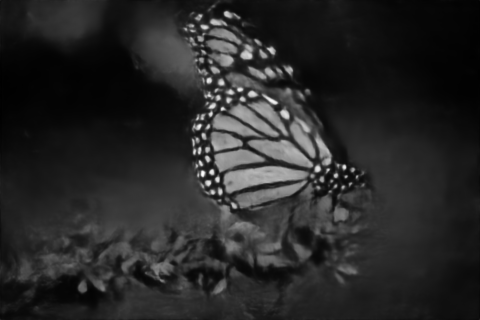}} %
	\subfigure[]{\includegraphics[width=0.32\linewidth]{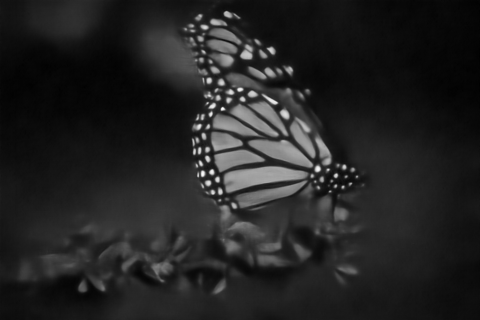}}
	\caption{Denoising result at additive noise of 50 standard deviation, of the proposed multi-scale network trained by our edge preservation loss. (a) The noisy input image. (b) The results of the proposed scheme. (c) Denoising results of the state-of-the-art DnCNN \protect\cite{zhang2017beyond} approach. Our method achieves the highest SSIM \protect\cite{ssim} scores in our experiments at all the noise levels.}
	\label{fig:butterfly}
\end{figure}

Edge detection is one of the fundamental problems of computer vision. Many
works addressed this problem and introduced a variety of solutions.
Unfortunately, some imaging domains suffer from faint edges and noisy
images, such as medical, satellite and even real natural images. The
detection of edges under such challenging conditions should be implemented
by methods geared to that end. Existing approaches that deal with high level
of noise are all relatively slow (runtime of seconds for an image).

This work is the first to use deep learning to detect faint edges, denoted as
Faint Edges Detection CNN (FED). By training the FED \cite{unet} on a
simulated faint-edges dataset, we developed a novel approach for edge
detection in noisy images. Since a forward pass of a network can be
optimized on a GPU, our algorithm is real time and is orders of magnitude
faster than existing approaches. Denote by $N$ the image pixels, than our
runtime is $O(N)$, while the FastEdges state-of-the-art approach \cite%
{ofir2016fast} runs in $O(NlogN)$. Even though, our experiments demonstrate
that it is yet more accurate in the task of binary edge detection at low
Signal-to-Noise-Ratios (SNR), where the SNR is the ratio of the edge
contrast $c$ to the noise level $\sigma $, such that $SNR=\frac{c}{\sigma }$%
. Edge detection of a medical noisy image is depicted in Fig. \ref{fig:1}.

The similarity between the classical vision approach of FastEdges \cite%
{ofir2016fast} and our method DeepFaster, is that both tackle noisy edge
detection by utilizing multi-scale preprocessing of the image. Ofir et al.
\cite{ofir2016fast} utilize a binary partitioning tree of the image, into
sub areas, and compute edge-filter responses at every sub-rectangle of image
pixels and concatenate curves from each sub-rectangle using dynamic
programming like approach. The proposed denoising scheme aims to mimic their
multi-scale filters using a convolutional neural network (CNN). We show that
the multi-scale processing of the image, using a CNN can be carried out by
the U-Net architecture. In addition to faint edge detection, we show that
our approach can be applied to additional computer vision tasks, such as to
improves the performance of a classifier trained for noisy image
classification. Specifically, let $I$ and $\widehat{I}$ be an input image
and its corresponding computed edge map, respectively, we show that the
accuracy of the Resnet20 \cite{resnet} classifier, on a noisy CIFAR10 \cite%
{cifar} dataset, is increased. This emphasizes the importance of multi-scale
filters and preprocessing to perform vision tasks at low SNR's.

We also apply an edge detector as an auxiliary loss for image
denoising. We train U-Net to perform denoising, and by that developing a
deep-multi-scale algorithm in a state-of-the-art level. We use for training
a novel architecture that utilizes an edge preservation auxiliary loss. Our
results of denoising are excellent in the perceptual measurement of
Structure-of-Similarity (SSIM) and Peak-Signal-to-Noise-Ratio (PSNR). See
Figure \ref{fig:butterfly} for example of our denoising approach relative to
the state-of-the-art DnCNN \cite{zhang2017beyond}. We managed to remove the
noise and preserve the signals in the highest quality among the existing
denoising algorithms.

\section{Previous Work}

\label{sec:previous}

Edge detection is a fundamental problem in image processing and computer
vision with a plethora of related works. Marr and Hildreth \cite%
{marr1980theory} studied edge detection using the zero crossing of the 2D
Laplacian applied to an image, while Sobel \cite{sobel} proposed to applying
a $3\times 3$ derivative filter on an image, and computing the gradients.
Canny \cite{canny1987computational} extends Sobel by hysteresis thresholding
of the local gradients. These classical approaches are very fast, but
unfortunately very sensitive to image noise and cannot accurately detect
faint edges. Advanced group of works are focused on the problem of boundary
detection and segmentation \cite{mcg,crisp,dollar2013structured}, achieving
accurate results when applied to the Berkeley Segmentations Dataset
(BSDS500) \cite{unnikrishnan2005measure}. A recent class of works are
optimized and trained on this dataset utilizing deep learning tools \cite%
{cob,hed,cedn}. Even though such approaches perform well for boundary
detection, their accuracy degrades in the presence of noise as was shown by
Ofir at al. \cite{ofir2016fast}.

The particular problem of faint edge detection in noisy images was
addressed by Galun et al. \cite{galun2007multiscale} by detecting faint edges
using the difference of oriented means. They applied a matched filter that
averages along each side of the edge, and maximizes the contrast across the
edge. Their method is limited to straight lines, with a computational
complexity of $O(NlogN)$ where $N$ is the number of pixels in an image. Ofir
et al. \cite{ofir2016fast} extended this work to curved edges utilizing
dynamic programming and approximations, to achieves better accuracy at a
complexity of $O(NlogN)$. In practice, the run time of these methods on a
noisy image is seconds. Sub-linear approaches \cite%
{horev2015detection,wang2017detecting} were introduced for detecting
straight and curved edges.

The proposed scheme utilizes deep-learning to improve these results, and we
introduce a $O(N)$ algorithm, whose actual runtime is negligible due to GPU
acceleration. Although our method is faster, we achieve even more accurate
results when detecting faint edges in noisy images. We utilize the U-Net
architecture \cite{unet} that was first derived for biomedical image
segmentation. We show that due to its multi-scale processing, it allows to
perform other vision tasks under hard conditions of low signals and high
noise.

Classification is a fundamental task in machine learning and computer
vision. Early methods to train a classifiers utilized
Support-Vector-Machines (SVM) \cite{svm} and logistic-regression \cite%
{logistic}. Deep learning approaches proved superior in classification
accuracy. The Resnet CNN architecture \cite{resnet} emphasized the
importance of residual connections in classification neural networks. These
networks are also the foundation of object detection and localization as
described in Single-Shot-Multibox-Detector (SSD) \cite{ssd}. All the above
approaches suffer from the presence of noise and objects at low SNR's. We
use as an example the Resnet20 \cite{resnet} classifier and CIFAR10 dataset
\cite{cifar} to exemplify the importance of a multi-scale preprocessing of
the noisy image to produce clean heat-maps. As done with faint edges, we use
the U-Net \cite{unet} architecture for this preprocessing.

Image denoising is one of the most studied areas of image processing and
computer vision. Early methods rely on the Wavelet Transform \cite%
{wavelet_denoising}. Advanced methods are based on patch repetitions in the
image like Non-Local-Means (NLM) \cite{nlm} and Block-Matching and
3D-filtering (BM3D) \cite{bm3d}. Recent methods utilize CNNs for denoising
\cite{zhang2017beyond,schmidt2014shrinkage}. In this work we show that the
multi-scale processing of U-Net \cite{unet} is useful for handling noisy
images, and to produce competitive denoising results on noisy natural
images. It is carried out by a novel architecture that utilizes an edge
preservation as an auxiliary loss.

\section{Faint Edges Detection in Noisy Images}

\label{sec:learning}

\begin{figure}[tbh]
\centering\includegraphics[width=220px]{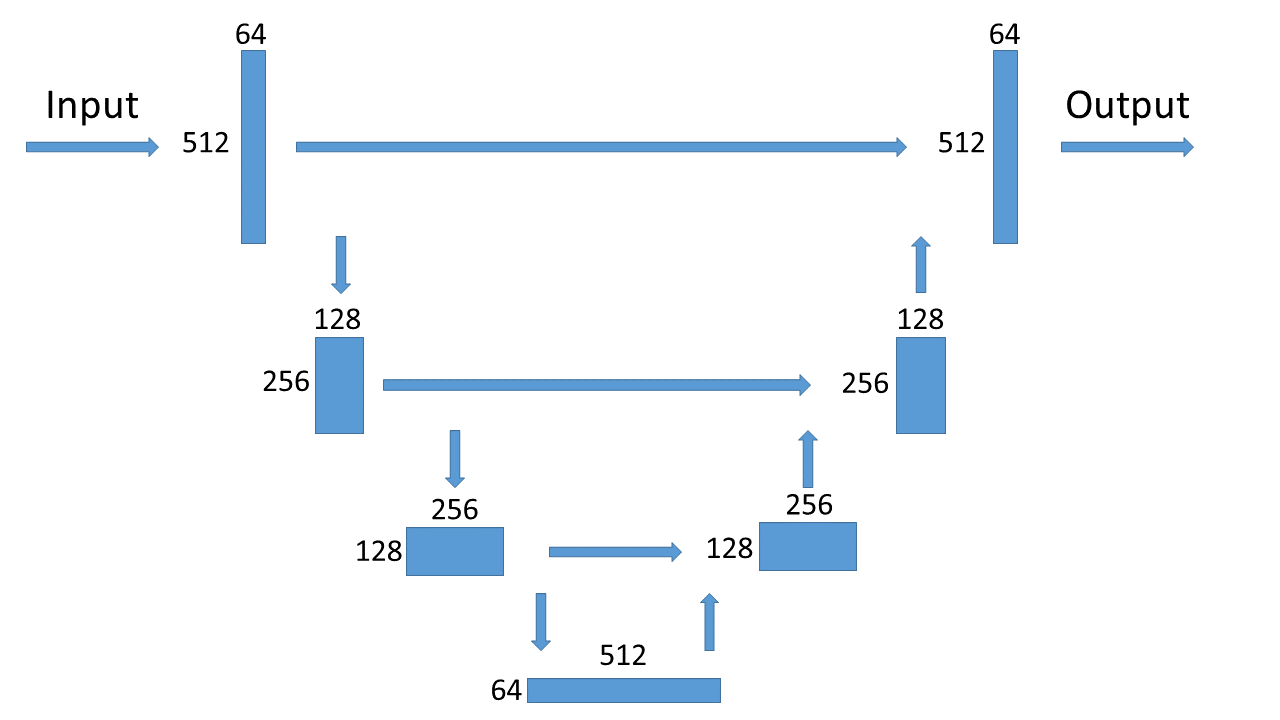}
\caption{The multi-scale architecture of U-net used in this work. The
network downscaled the activation map by a series of max-pooling layers in
the first half, and upsampled by interpolation layers in its second half.
There are concatenation connections between the first half and the second
half for every two layers having the same number of channels.}
\label{fig:unet}
\end{figure}

We propose the Faint Edges Detection CNN (FED) to formulate the detection of
faint edges in noisy images as a binary image segmentation, based on the
U-Net \cite{unet} architecture that is depicted in Fig. \ref{fig:unet}. This
FED allows to create a multi-scale algorithms for detection and
segmentation, where sigmoid activations are appended on top of the last
layer. The proposed FED CNN was trained using the Dice coefficients. Denote
by $y^{\prime }$ the binary labels of the edges, and by $y$ the output of
the network for a given input $x$, the Dice coefficient is given by%
\begin{equation}
Di(y,y^{\prime })=-\frac{\sum_{p}y^{\prime }(p)\cdot y(p)}{\sum_{p}y^{\prime
}(p)+\sum_{p}y(p)},  \label{eq:dice}
\end{equation}%
where $p$ is an image pixel. This loss encourages accurate detections, while
penalizes false alarms.

We initialize the FED using random weights, and used a dataset of 1406
binary images to train the network. For each ground truth image we apply a
Canny edge detector \cite{canny1987computational} to extract the labels $%
Y^{\prime }$, and used to create a set of noisy images having different
signal to noise ratios (SNR's)
\begin{equation}
I=clip(0.1\ast (snr\cdot I_{c}+I_{n})+0.45).  \label{eq:noise}
\end{equation}%
$I$ is the noisy image, input to our network, $I_{c}$ is the original binary
image, $I_{n}$ is a random Gaussian noise image with standard deviation of
1, $\forall p:I_{n}(p)\sim N(0,1)$. The $snr$ is the measure of the
faintness of the edges, each binary image creates six training images $snr=%
\left[ 1,1.2,...,2\right] $. The $clip\left( \cdot \right) $ clips the pixel
values to $\left[ 0,1\right] $.

We trained the FED for 100 epochs and augmented the dataset by different
snr's and horizontal and vertical flipping of the images. Moreover, we
added samples of a pure noise image with no labels. Our dataset after
augmentations contains $\sim $17,000 examples. We split the dataset to $90\%$
training and $10\%$ testing.
\begin{table}[tbh]
\centering%
\begin{tabular}{|c|c|c|c|c|c|}
\hline
\# & Type & Out Dim & Kernel & Stride & Pad \\ \hline\hline
1 & convolution & 64 & 3$\times $3 & 1 & 1 \\ \hline
2 & ReLU & 64 & - & 1 & 0 \\ \hline
3 & convolution & 64 & 3$\times $3 & 1 & 1 \\ \hline
4 & ReLU & 64 & - & 1 & 0 \\ \hline
5 & max-pooling & 64 & 2$\times $2 & 2 & 0 \\ \hline
6 & convolution & 128 & 3$\times $3 & 1 & 1 \\ \hline
7 & ReLU & 128 & - & 1 & 0 \\ \hline
8 & convolution & 128 & 3$\times $3 & 1 & 1 \\ \hline
9 & ReLU & 128 & - & 1 & 0 \\ \hline
10 & max-pooling & 128 & 2$\times $2 & 2 & 0 \\ \hline
11 & convolution & 256 & 3$\times $3 & 1 & 1 \\ \hline
12 & ReLU & 256 & - & 1 & 0 \\ \hline
13 & convolution & 256 & 3$\times $3 & 1 & 1 \\ \hline
14 & ReLU & 256 & - & 1 & 0 \\ \hline
15 & max-pooling & 256 & 2$\times $2 & 2 & 0 \\ \hline
16 & convolution & 512 & 3$\times $3 & 1 & 1 \\ \hline
17 & ReLU & 512 & - & 1 & 0 \\ \hline
18 & convolution & 512 & 3$\times $3 & 1 & 1 \\ \hline
19 & ReLU & 512 & - & 1 & 0 \\ \hline
20 & UpSample & 512 & - & 0.5 & 0 \\ \hline
21 & cat(20,14) & 768 & - & - & 0 \\ \hline
22 & convolution & 256 & 3$\times $3 & 1 & 1 \\ \hline
23 & ReLU & 256 & - & 1 & 0 \\ \hline
24 & convolution & 256 & 3$\times $3 & 1 & 1 \\ \hline
25 & ReLU & 256 & - & 1 & 0 \\ \hline
26 & UpSample & 256 & - & 0.5 & 0 \\ \hline
27 & cat(26,9) & 384 & - & - & 0 \\ \hline
28 & convolution & 128 & 3$\times $3 & 1 & 1 \\ \hline
29 & ReLU & 128 & - & 1 & 0 \\ \hline
30 & convolution & 128 & 3$\times $3 & 1 & 1 \\ \hline
31 & ReLU & 128 & - & 1 & 0 \\ \hline
32 & UpSample & 128 & - & 0.5 & 0 \\ \hline
33 & cat(32,4) & 192 & - & - & 0 \\ \hline
34 & convolution & 64 & 3$\times $3 & 1 & 1 \\ \hline
35 & ReLU & 64 & - & 1 & 0 \\ \hline
36 & convolution & 64 & 3$\times $3 & 1 & 1 \\ \hline
37 & ReLU & 64 & - & 1 & 0 \\ \hline
38 & convolution & 1 & 1$\times $1 & 1 & 0 \\ \hline
39 & Sigmoid & 1 & - & 1 & 0 \\ \hline
\end{tabular}%
\caption{The multi-scale architecture of U-net. The network has a 'U' shape,
it downscale the image by a series of max-pooling layers in the first part,
and up sample by interpolation layers in the second part. There are
concatenation connections between the first part to the second for every two
layers in the same image dimensions. Due to its architecture, the network
applies multi-scale filters that maximize the signals and average the noise.
Therefore it allows detection of edges and objects at low SNR's.}
\end{table}

\section{Classification of Noisy Images}

\label{sec:classifier}

In this Section, we apply the proposed FED to the classification of noisy
images, using a Resnet20 CNN \cite{resnet}. Denote by $x$ an input images,
in CIFAR10 dataset \cite{cifar} $x_{i}\in
%TCIMACRO{\U{211d} }%
%BeginExpansion
\mathbb{R}
%EndExpansion
^{32\times 32\times 3}$. Given the Resnet classifier, the label is given by $%
y_{i}=resnet(x_{i})$. We trained this model using the CIFAR10 dataset and
achieved classification accuracy of $91.66$\%. Denote the Resnet network
trained on the regular clean CIFAR10 as $resnet_{c}$. We created a noisy
version of CIFAR10, by adding Gaussian noise of different standard
deviations to the image. The accuracy of $resnet_{c}$ on
noisy CIFAR10 is only $34.1$\%. This accuracy shows that the accuracy of
conventional classification CNNs, that achieve excellent results when applied
to clean images, dramatically degrade due to the presence of noise.

\begin{figure}[tbh]
\centering
\fbox{\includegraphics[width=220px]{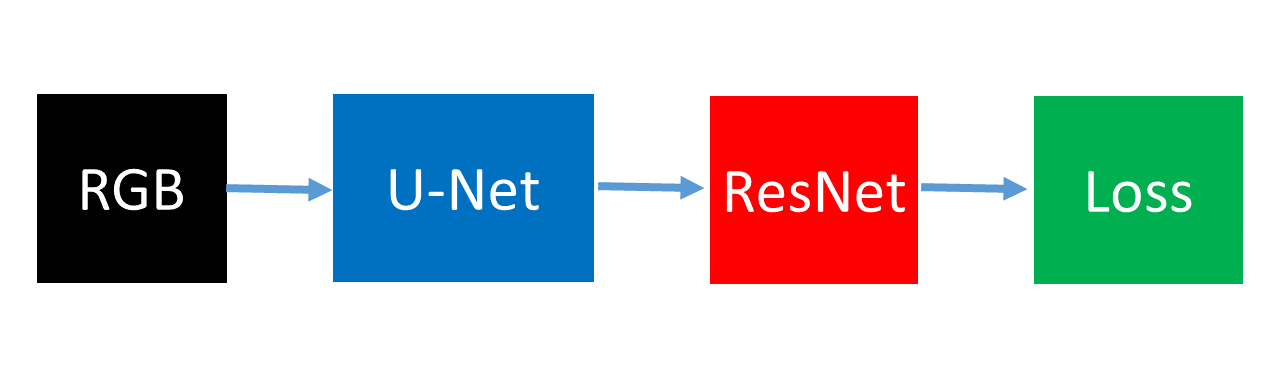}}
\caption{Visualization of our classification process. The input is an RGB
image, typically with high level of noise. Then, it is processed by the
multi-scale filters of U-Net to produce meaningful heat-maps. Resnet 20
classifier process the heat-map and outputs the classification label.
Finally, the whole scheme is trained using classification cross-entropy
loss. }
\label{fig:process}
\end{figure}

The straight-forward approach is to train Resnet using a noisy dataset,
denote this network by $resnet_{n}$. The accuracy of this network on noisy
CIFAR10 increases significantly to $77.5$\%. However, we aim to filter out
the noise using the proposed FED CNN. Given a noisy image $x_{i}$, our
preprocessing will produce a noise free heat map for classification purposes
such that $h=FED(x_{i})$. Then we classify with Resnet using this heatmap
such that $y_{i}=resnet(FED(x_{i}))$. Note the this replica of Resnet is
trained using the noisy CIFAR10, and we train end-to-end, such that the
heat-maps for classification might differ from those for edge-detection.
Figure \ref{fig:process} shows the outline of the proposed classification
process.

\section{Image Denoising using an Edges-Guided Auxiliary Loss}

\label{sec:denoising}
\begin{figure}[tbh]
\centering
\fbox{\includegraphics[width=220px]{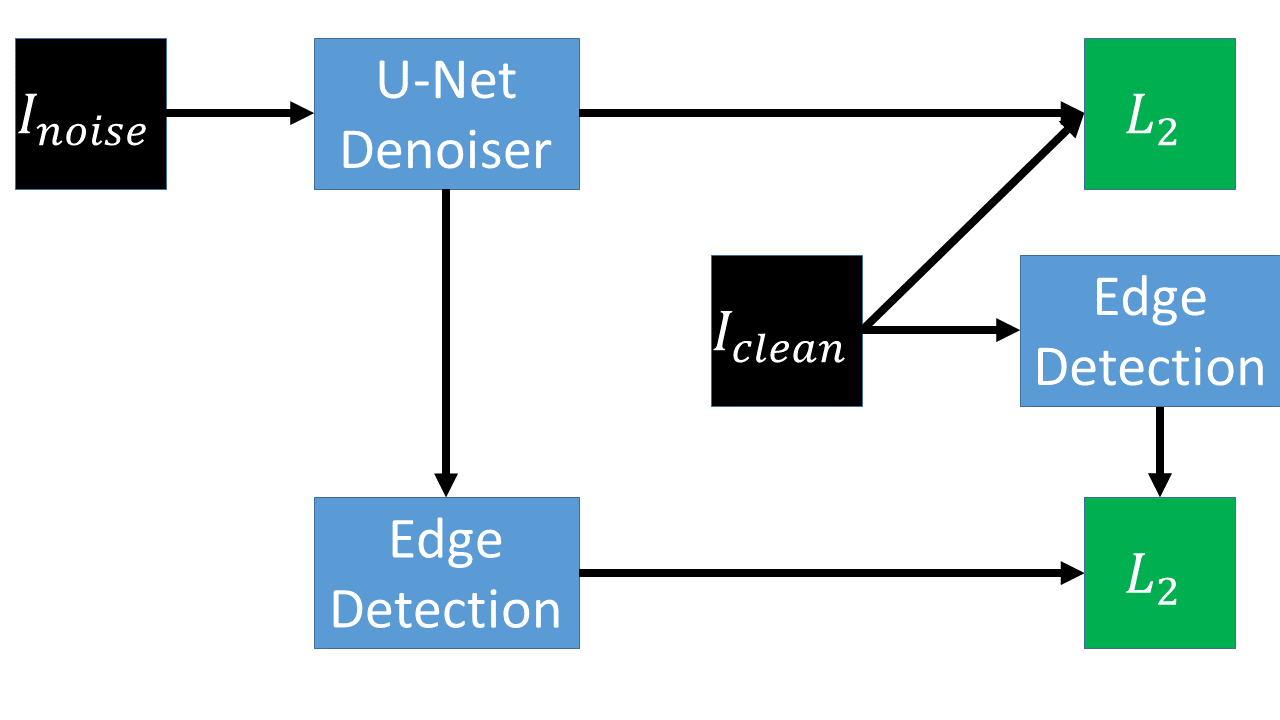}}
\caption{Architecture of our training of U-Net for natural image denoising.
The input noisy image $I_{noise}$ is the input for the U-Net denoising. Then,
the output is connected to $L_{2}$ loss compared to the ground truth $%
I_{clean}$. In addition, we add an edge preserving auxiliary loss. By
another $L_{2}$ loss we compare the Sobel edge detection \protect\cite{sobel}
of $I_{clean}$ to the edges map of the denoised image: $%
Edges(IDCNN(I_{noise})) $. This scheme improve the performance of our
denoising over the regular $L_{2}$ loss training in both aspects of PSNR and
SSIM.}
\label{fig:auxiliary}
\end{figure}

In this section we continue to exemplify the use of edge detection in image denoising.
We introduce a novel approach to train an image denoising with an auxiliary
loss of edge preservation. To that end, we train an Image Denoising CNN
(IDCNN) based on the U-Net architecture for images denoising. We use the
images in the Berkeley-Segmentation-Dataset (BSDS), where each image is first
converted to gray-scale, to create the pairs $\left\{ I_{c},I_{n}\right\} $
of clean and corresponding noisy images. The noisy images were computed by
adding white Gaussian noise to each pixel using the noise $n\sim N(0,\sigma^2
=15^2,25^2,50^2)$.

The input for training IDCNN is the noisy image $I_{n}$ and the output is
trained to be as closest to $I_{c}$ using a $L_{2}$ regression loss. The
IDCNN is trained for 200 epochs. To improve the quality of the results we
train the same architecture called IDCNN by and edge preservation auxiliary loss, and the resulting
architecture is depicted in Fig. \ref{fig:auxiliary}. The overall loss is
the sum of the $L_{2}$ loss and the edges loss. The label for the loss is
the Sobel edge detector \cite{sobel} applied to the clean image. The labels
are compared to the result of the same edge maps of IDDCNN%
\begin{equation}
L_{E}=||\frac{\partial }{\partial x}I_{c}-\frac{\partial }{\partial x}%
IDCNN(I_{n})||_{2}^{2}.
\end{equation}

Our experiments show that our auxiliary loss improve the performance of the
denoising PSNR and SSIM. In comparison to the excellent deep approach
of DnCNN \cite{zhang2017beyond}, we achieve state-of-the-art level of image
denoising. We show that the proposed IDCNN-Edges (IDCNN-E) is able to denoise the image while preserving the image details and it is the best approach for maximizing the SSIM score of the denoising result.

\section{Experimental Results}

\label{sec:experimetns}

We experimentally verified the proposed scheme using simulated and real
images. In edge detection we compare to FastEdges \cite{ofir2016fast},
classic Canny edge detector \cite{canny1987computational} and to the
CNN-based Holistically-Edge-Detection (HED) \cite{hed}. We test these
methods in both aspects of quality and run time. To extract a measure of
similarity between the binary results and the ground truth we use a strict
version of F-measure, that applies pixel-wise accuracy and does not allow
the matching of neighboring pixels. The F-measure is the harmonic mean of
the precision and recall:
\begin{equation}
F=\frac{2\cdot precision\cdot recall}{precision+recall}.
\end{equation}%
Denote by $Y^{\prime }$ the labels and by $Y$ the results, the precision is
\begin{equation}
precision=\frac{TruePositive}{TruePositive+FalsePositive}=\frac{\sum Y\cdot
Y^{\prime }}{\sum Y},
\end{equation}%
while the recall is given by%
\begin{equation}
recall=\frac{TruePositive}{TruePositive+FalseNegative}=\frac{\sum Y\cdot
Y^{\prime }}{\sum Y^{\prime }}.
\end{equation}

\begin{figure}[tbh]
\centering
\fbox{\includegraphics[width=60px]{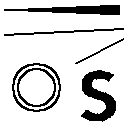}}
\caption{Simulation of faint edges detection in noisy image. The binary
pattern was used to evaluate the performance of the methods in Figure \protect
\ref{fig:4}}
\label{fig:binaryPattern}
\end{figure}

\begin{figure}[tbh]
\centering
\includegraphics[width=200px]{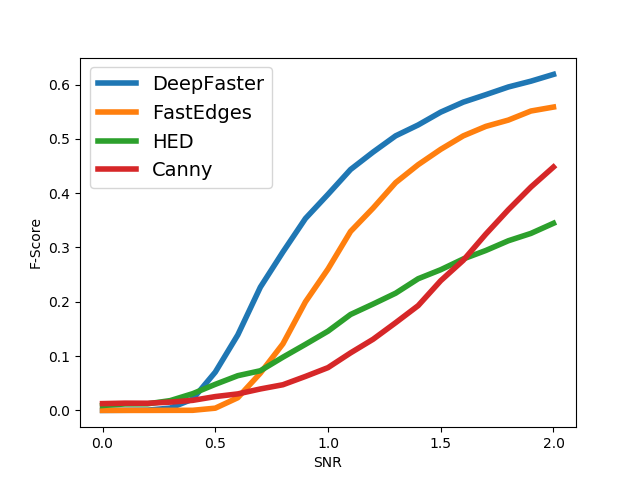}~
\caption{Simulation of faint edges detection in noisy image. The strict
F-score graph along the different signal-to-noise ratios from 0 to 2. The
methods geared for faint edges detection achieve the highest accuracy. Our
method DeepFaster obtains a slightly higher graph from the previous approach
of FastEdges \protect\cite{ofir2016fast}.}
\label{fig:4}
\end{figure}

We compared the algorithms using a challenging binary patterns shown in
Fig. \ref{fig:binaryPattern}. This pattern contains triangle, straight
lines, 'S' shape and concentric circles. We used it to create a set of noisy
images with $SNR=\left[ 1,1.2,...,2\right] $ Gaussian additive noise. For
each SNR, we compute the F-measure for every method for 100 iterations of
different random noises as in Eq. \ref{eq:noise}. Then we take the average
F-score of all the iterations. It can be seen in Fig. \ref{fig:4} that the
graph of the methods geared to detect faint edges in noisy images, ours and
\cite{ofir2016fast}, are superior to Canny \cite{canny1987computational}.
Moreover, the proposed FEDCNN outperforms the FastEdges approach.

\begin{figure}[tbh]
\centering
\fbox{\includegraphics[width=70px]{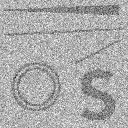}}~ \fbox{%
\includegraphics[width=70px]{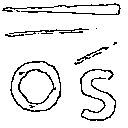}}~ \fbox{%
\includegraphics[width=70px]{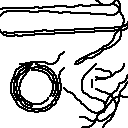}}
\caption{Results on the simulation noisy image. Left: the original images at
SNR=2. Middle: our network result. Right: FastEdges \protect\cite%
{ofir2016fast} result. Both methods produce good quality results while ours
contains less false detections.}
\label{fig:5}
\end{figure}

The results are summarized in Table \ref{table:1}, and it follows that for $%
SNR=\left\{ 1,2\right\} $, the proposed FEDCNN approach achieves the highest
F-score. Figure \ref{fig:5} shows the simulation images of $SNR=2$. Compared
to FastEdges \cite{ofir2016fast}, both approaches achieve a good accuracy of
detection on this image, but the FEDCNN yields less false detections.

Even though the FEDCNN is the most accurate, it achieves a real-time running
time as shown in Table \ref{table:2}. Its runtime is 10 millisecond, similar
to Canny's run time, and it is orders of magnitude faster that FastEdges
\cite{ofir2016fast} that runs in seconds. We computed these run times on a
single machine with i7 CPU, 32 GB of RAM, and geforce gtx 1070 GPU. Note
that the Canny and FastEdges implementations run on the CPU while our new
method utilized the parallelism of the GPU.
\begin{table}[tbh]
\centering%
\begin{tabular}{|l|c|c|}
\hline
Algorithm & SNR = 1 & SNR = 2 \\ \hline\hline
DeepFaster & \textbf{0.4} & \textbf{0.62} \\ \hline
FastEdges & 0.28 & 0.56 \\ \hline
HED & 0.14 & 0.34 \\ \hline
Canny & 0.08 & 0.45 \\ \hline
\end{tabular}%
\caption{F-score of the methods at SNR 1 and 2. At both snr's our method
achieves the highest score.}
\label{table:1}
\end{table}
\begin{table}[tbh]
\centering%
\begin{tabular}{|l|c|}
\hline
Algorithm & Run-Time (milliseconds) \\ \hline\hline
DeepFaster & \textbf{10} \\ \hline
FastEdges & 2600 \\ \hline
Canny & 3 \\ \hline
\end{tabular}%
\caption{Run time in milli-seconds of the different methods of edge
detection. Our runtime is very close to Canny's time and is order of
magnitude faster than FastEdges. We achieve this improvement by running our
network on a GPU.}
\label{table:2}
\end{table}

\begin{figure}[tbh]
\centering
\fbox{\includegraphics[width=70px]{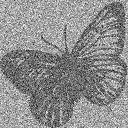}}~ \fbox{%
\includegraphics[width=70px]{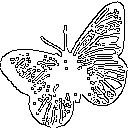}}~ \fbox{%
\includegraphics[width=70px]{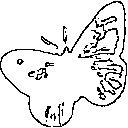}}\\[0.1cm]
\fbox{\includegraphics[width=70px]{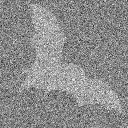}}~ \fbox{%
\includegraphics[width=70px]{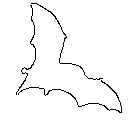}}~ \fbox{%
\includegraphics[width=70px]{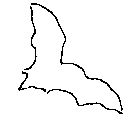}}\\[0.1cm]
\fbox{\includegraphics[width=70px]{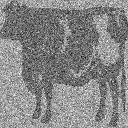}}~ \fbox{%
\includegraphics[width=70px]{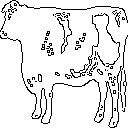}}~ \fbox{%
\includegraphics[width=70px]{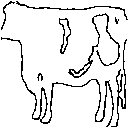}}\\[0.1cm]
\fbox{\includegraphics[width=70px]{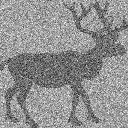}}~ \fbox{%
\includegraphics[width=70px]{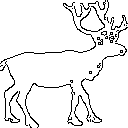}}~ \fbox{%
\includegraphics[width=70px]{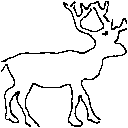}}
\caption{Results on images from the binary images dataset that we used to train and test our
network. Left: the input noisy images with a binary pattern. Middle: the
ground truth labels. Right: our detections. DeepFaster result is very
similar to the ground truth and we manage to detect and track edges even at
high curvatures.}
\label{fig:6}
\end{figure}

\begin{figure}[tbh]
\centering
\fbox{\includegraphics[width=70px]{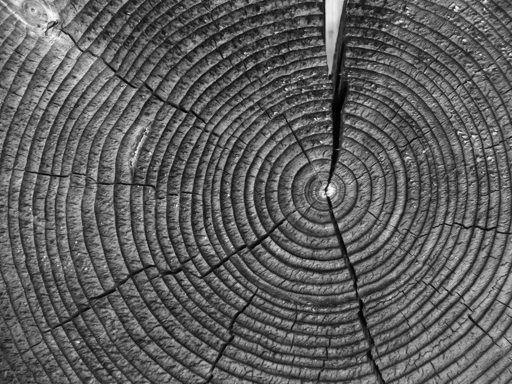}}~ \fbox{%
\includegraphics[width=70px]{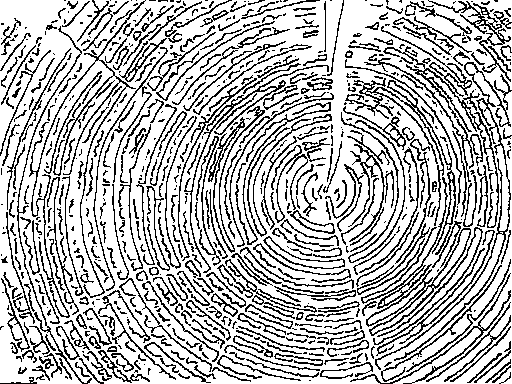}}~ \fbox{%
\includegraphics[width=70px]{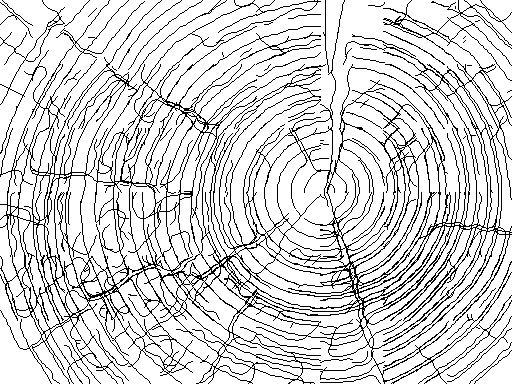}}\\[0.1cm]
\fbox{\includegraphics[width=70px]{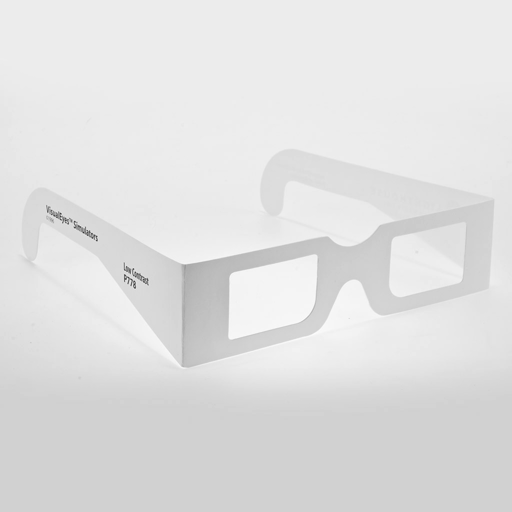}}~ \fbox{%
\includegraphics[width=70px]{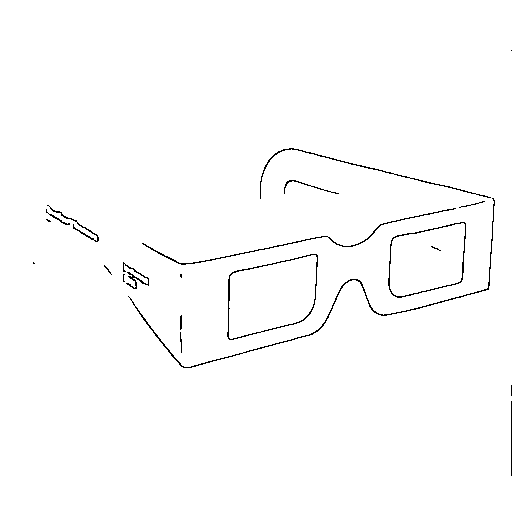}}~ \fbox{%
\includegraphics[width=70px]{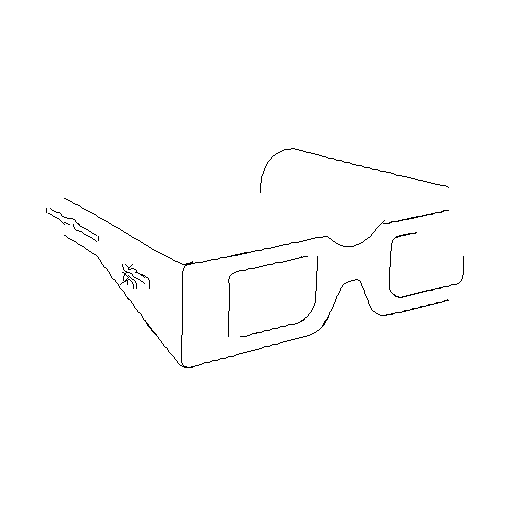}}\\[0.1cm]
\fbox{\includegraphics[width=70px]{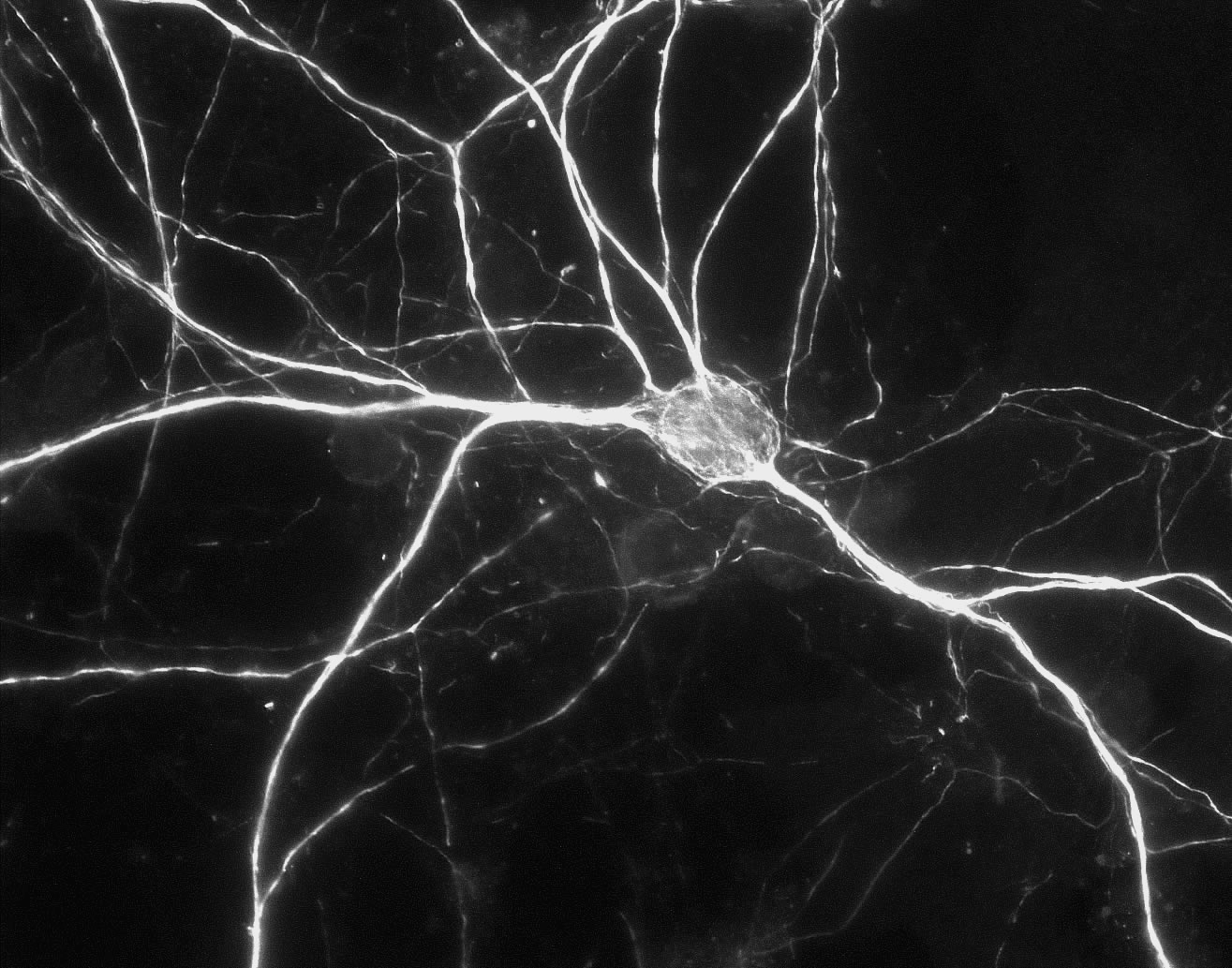}}~ \fbox{%
\includegraphics[width=70px]{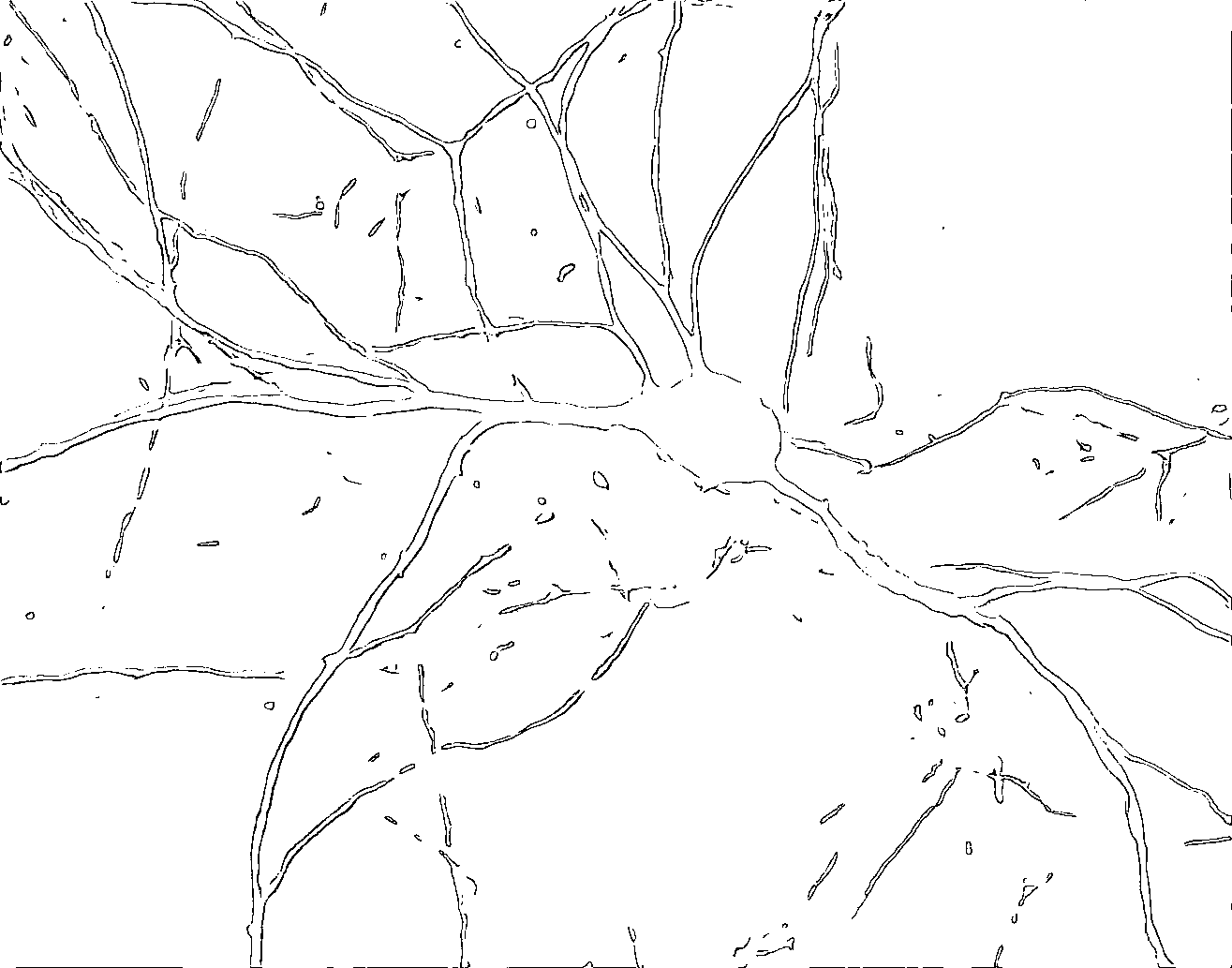}}~ \fbox{%
\includegraphics[width=70px]{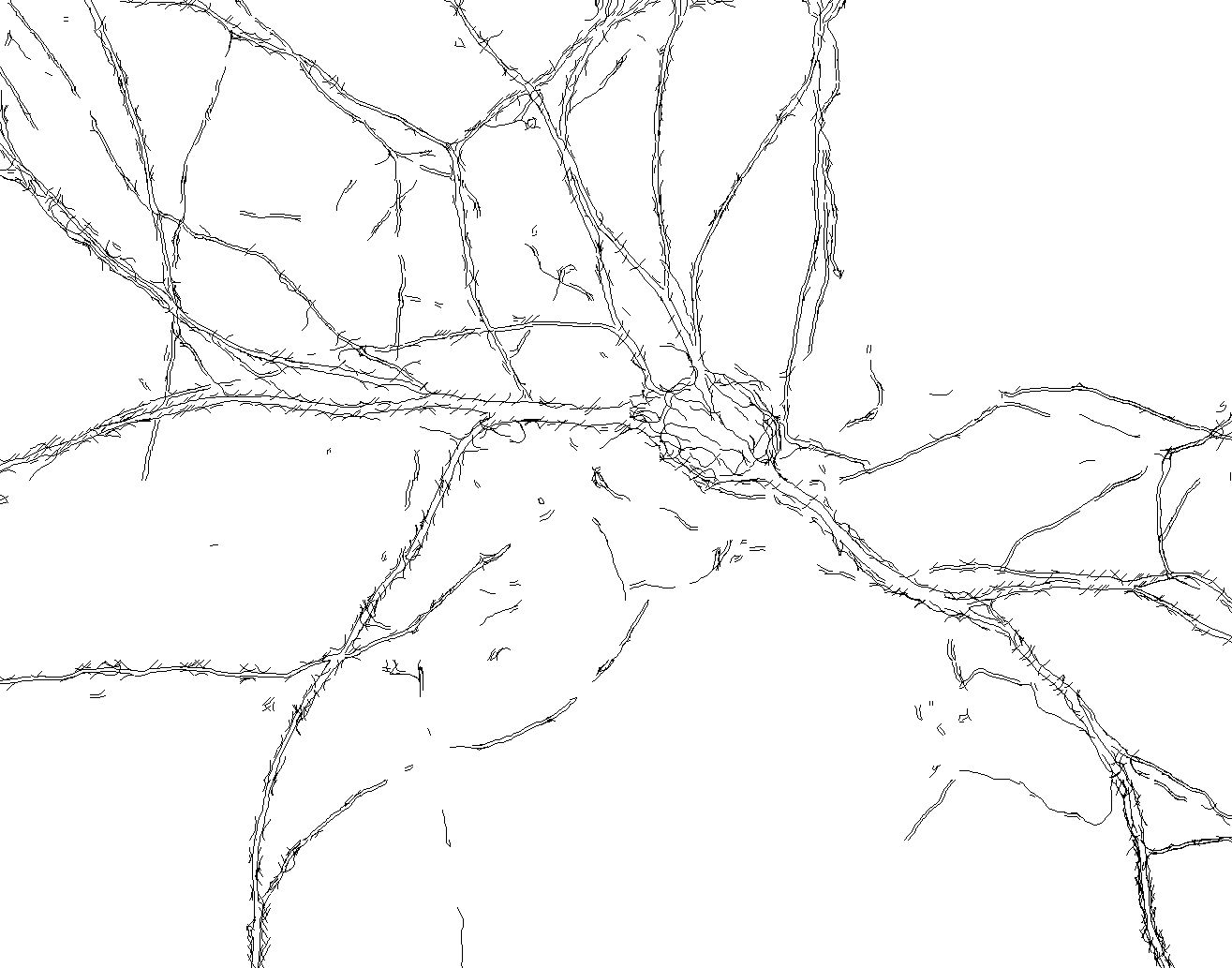}}\\[0.1cm]
\fbox{\includegraphics[width=70px]{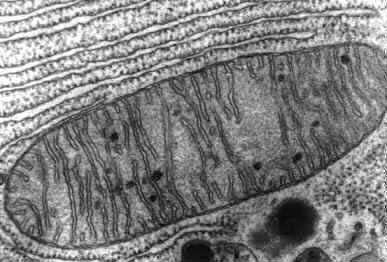}}~ \fbox{%
\includegraphics[width=70px]{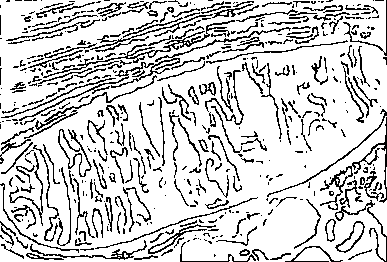}}~ \fbox{%
\includegraphics[width=70px]{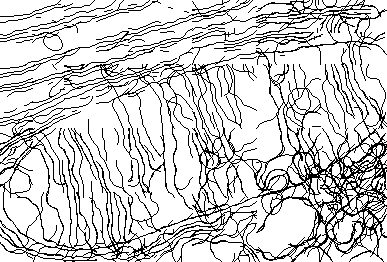}}\\[0.1cm]
\fbox{\includegraphics[width=70px]{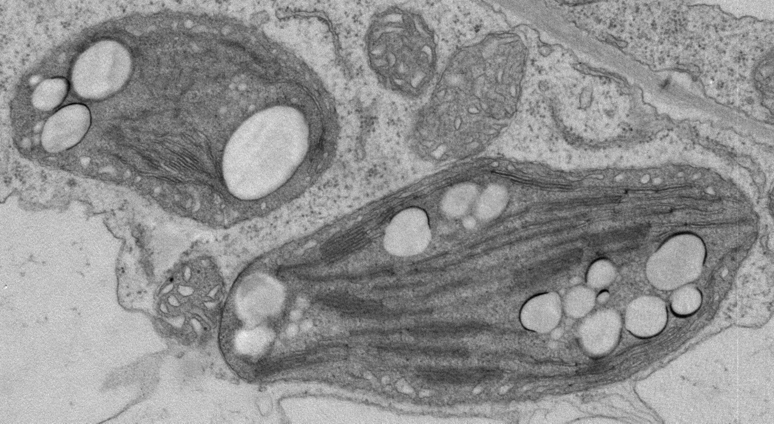}}~ \fbox{%
\includegraphics[width=70px]{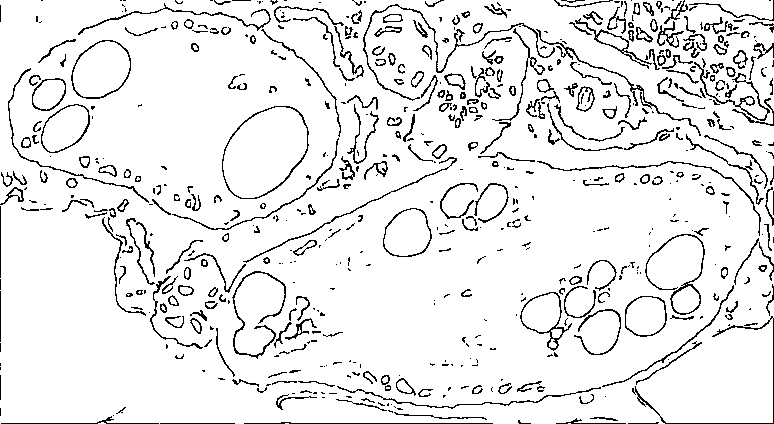}}~ \fbox{%
\includegraphics[width=70px]{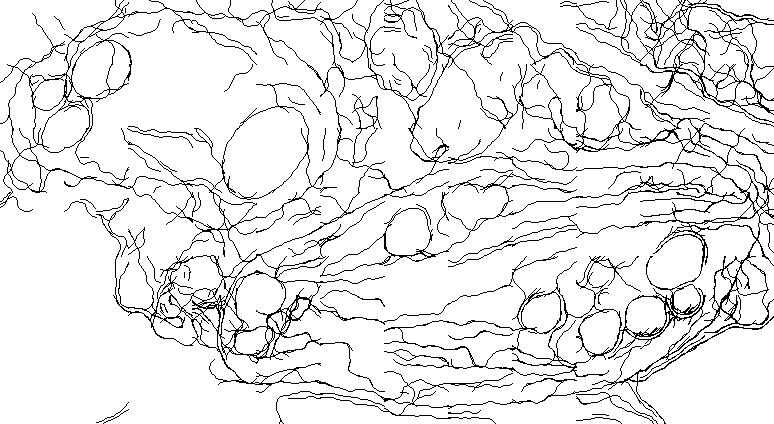}}
\caption{Examples of real images. Left: the original gray scale images.
Middle: our results. Right: FastEdges \protect\cite{ofir2016fast} results.
Both methods achieve high quality of detections.}
\label{fig:7}
\end{figure}

Figure \ref{fig:6} shows the FEDCNN results when applied to noisy images
from the binary images dataset used for training and testing. It follows that we
manage to detect and track high curvatures edges, being very similar to the
ground truth labels. Figure \ref{fig:7} show the FEDCNN and FastEdges \cite%
{ofir2016fast} results on a group of real images. Both methods obtains high
quality of detection. However, since our network is fully convolutional, its
run-time does not scale significantly with size, and our run time on these
images is much faster than FastEdges.

\subsection{Noisy Image Classification}

We also examined the applicability of the proposed FEDCNN to image
classification, following Section \ref{sec:classifier}. For that we studied
the following, we can classify images using the basic $resnet_{c}$, we can
train on noisy dataset and use $resnet_{noisy}$, or to add preprocessing by
U-Net and to use $resnet(FED)$.

\begin{figure}[tbh]
\centering
\includegraphics[width=0.9\linewidth]{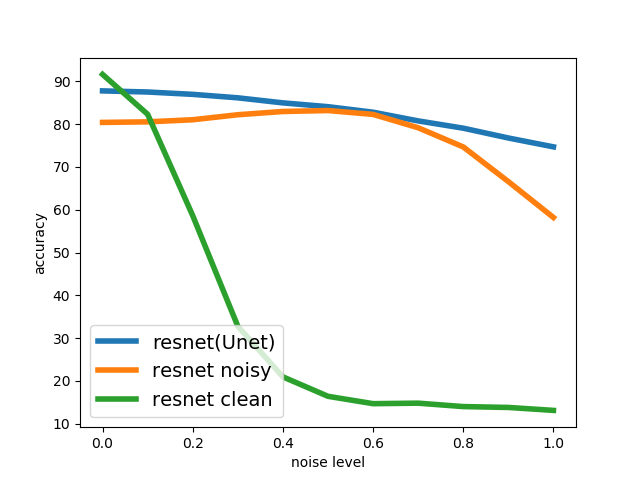}~
\caption{Accuracy of classifiers along different noise levels. Our approach,
training of U-net \protect\cite{unet} and resnet20 \protect\cite{resnet} on
a noisy dataset archives the highest accuracy at all levels greater than 0.
Training of resnet 20 on a noisy dataset, without the multi-scale
preprocessing of U-net, archives lower scores. The regular resnet 20
classifier, trained on clean dataset, achieves the highest score on noise
free images, but poor accuracy on all other noise levels.}
\label{fig:accuracyGraph}
\end{figure}

\begin{figure}[tbh]
\centering
\fbox{\includegraphics[width=70px]{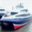}}~ \fbox{%
\includegraphics[width=70px]{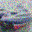}}~ \fbox{%
\includegraphics[width=70px]{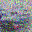}}\newline
\fbox{\includegraphics[width=70px]{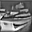}}~ \fbox{%
\includegraphics[width=70px]{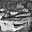}}~ \fbox{%
\includegraphics[width=70px]{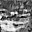}}\newline
\fbox{\includegraphics[width=70px]{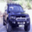}}~ \fbox{%
\includegraphics[width=70px]{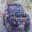}}~ \fbox{%
\includegraphics[width=70px]{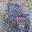}}\newline
\fbox{\includegraphics[width=70px]{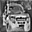}}~ \fbox{%
\includegraphics[width=70px]{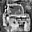}}~ \fbox{%
\includegraphics[width=70px]{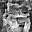}}\newline
\fbox{\includegraphics[width=70px]{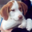}}~ \fbox{%
\includegraphics[width=70px]{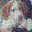}}~ \fbox{%
\includegraphics[width=70px]{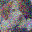}}\newline
\fbox{\includegraphics[width=70px]{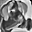}}~ \fbox{%
\includegraphics[width=70px]{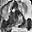}}~ \fbox{%
\includegraphics[width=70px]{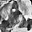}}\newline
\caption{Examples of noisy objects from the CIFAR 10 \protect\cite{cifar}
dataset and their heat-maps produced by U-net. Odd rows show the $32\times
32$ images of the object in 3 noisy levels: clean, moderate and high. Even rows show
the corresponding heat-map of the images produced my the multi-scale
preprocessing of FED architecture. These heat-maps allows classification
of objects under high levels of noise.}
\label{fig:heatmaps}
\end{figure}

We evaluated the performance of these three variants empirically. Table \ref%
{table:accuracy} shows the overall accuracy of classification to 10 or 100 classes. Our approach
is the only one that crosses the 80\% gap and achieves accuracy of $82.7$\% on noisy CIFAR10.
In addition, Figure \ref{fig:accuracyGraph} show the accuracy of each
approach at every noise level. The basic network of $resnet_{c}$ gains the
highest score of clean images, while our approach gets the highest scores
for all the other noise levels. We also emphasize our heat-map contribution
visually. Figure \ref{fig:heatmaps} show examples of images from the noisy
CIFAR10 test set. For each image we show the corresponding heat map
produced by FEDCNN. As can be seen, we succeed to produce meaningful
heat-maps for every noise level such that the structure of the object is
preserved and the noise is removed. As U-Net maximized the SNR for edges, it
does the same for maximizing objects visibility and SNR due to its multi-scale set of
filters.

\begin{figure}[tbh]
\centering
\fbox{\includegraphics[width=100px]{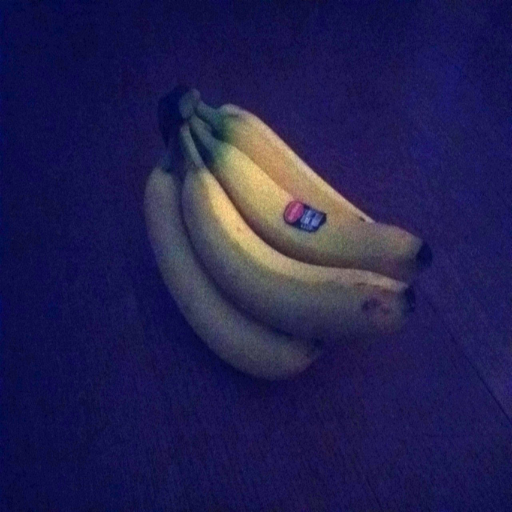}}~ \fbox{%
\includegraphics[width=100px]{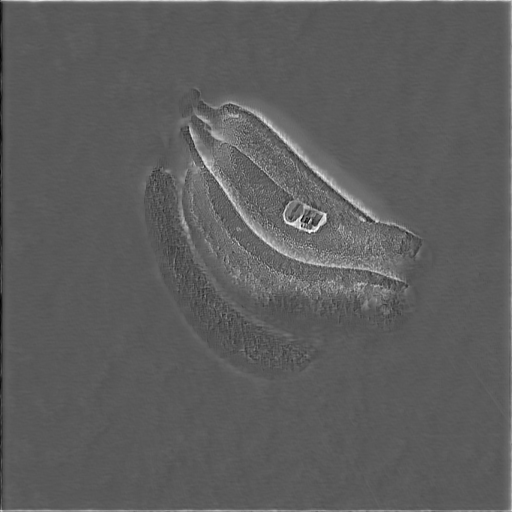}}~ \\[0.1cm]
\fbox{\includegraphics[width=100px]{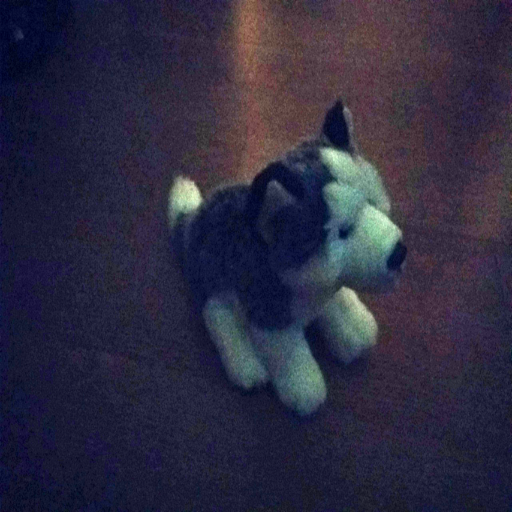}}~ \fbox{%
\includegraphics[width=100px]{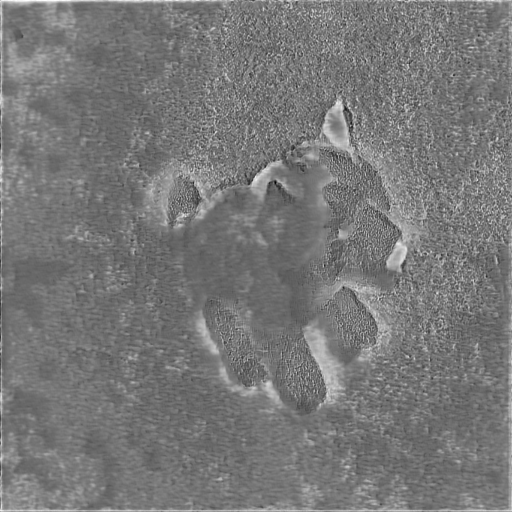}}~ \\[0.1cm]
%\fbox{\includegraphics[width=100px]{Images/lowlight0004_0.png}}~
%\fbox{\includegraphics[width=100px]{Images/lowlight0004_0_map.png}}~
%\\[0.1cm]
\fbox{\includegraphics[width=100px]{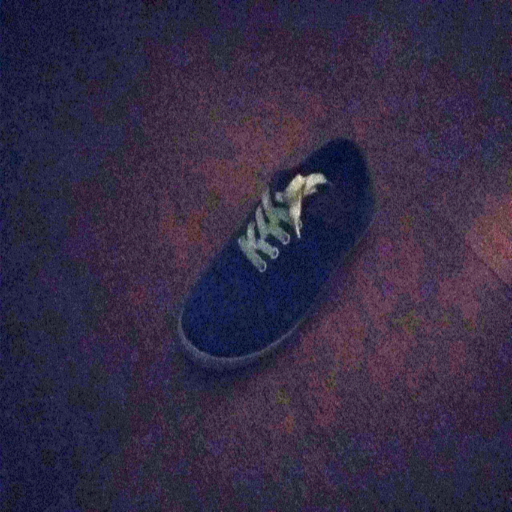}}~ \fbox{%
\includegraphics[width=100px]{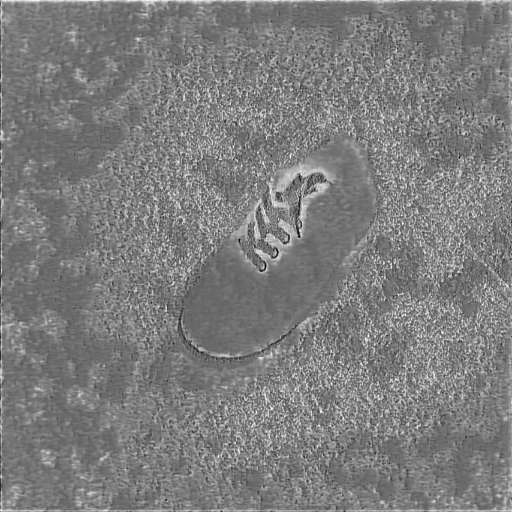}}~ \\[0.1cm]
\caption{Low light dark images and their heat-maps produced by FED. This
network was trained to output heat maps that will maximize the ability to
see and classify the captured objects. It can be seen that the object's
boundaries are emphasized although the low level of SNR.}
\label{fig:lowlight}
\end{figure}

We also examined our heat-maps for classification on low-light images in
Figure \ref{fig:lowlight}. The low light causes to low SNR image. Although
the high level of noise, our approach succeed to output meaningful heat maps
that emphasize the boundaries of the objects. These maps allows human or
machine vision of the dark objects.
\begin{table}[tbh]
\centering%
\begin{tabular}{|l|c|c|}
\hline
Algorithm & CIFAR10 & CIFAR100 \\ \hline\hline
$resnet(FED)$ & \textbf{82.7} & \textbf{53.3} \\ \hline
$resnet_{noisy}$ & 77.5 & 46.0 \\ \hline
$resnet_{c}$ & 34.1 & 16.9 \\ \hline
\end{tabular}%
\caption{Classification accuracy of the different methods and architectures
averaged on all noisy levels. Our approach of multi-scale
preprocessing by U-net \protect\cite{unnikrishnan2005measure} and
classification by resnet 20 \protect\cite{resnet} achieves the highest
accuracy.}
\label{table:accuracy}
\end{table}

\subsection{Image Denoising}

We evaluated the proposed IDCNN trained by $L_2$ regression loss and IDCNN-Edges (IDCNN-E) approach trained by an edge preservation auxiliary loss for image denoising using the
Berkeley-Segmentation-Dataset (BSDS) \cite{bsds}. Zero-mean Gaussian noise
with $\sigma =15,25,50$ was added to each image. We used the 200 train images to
train the IDCNN, and the 200 test images to evaluate our scheme. We compared
our approach to the classical methods of BM3D \cite{bm3d}
and to the state-of-the-art deep-approach DnCNN \cite{zhang2017beyond}. For
quantitative comparison we measured the Peak-Signal-to-Noise-Ratio (PSNR) and
Structure-of-Similarity (SSIM) \cite{ssim}. As shown in Table \ref%
{table:denoising}, the IDCNN-E achieves the highest perceptual SSIM score, and we
gain competitive PSNR. In particular, following \cite{blau2018perception},
there is a known perception-distortion trade-off, and it is difficult to
achieve both.

\begin{figure}[tbh]
\centering
\fbox{\includegraphics[width=70px]{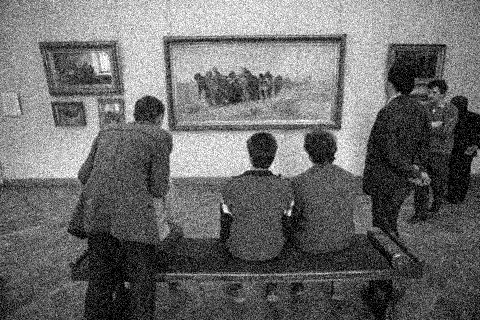}}~ \fbox{%
\includegraphics[width=70px]{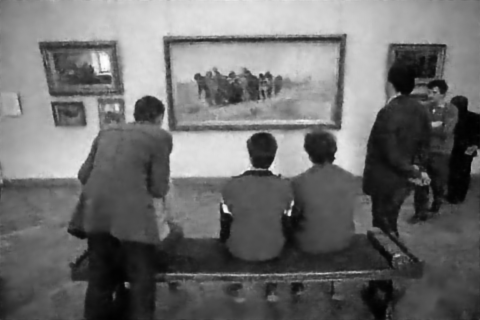}}~ \fbox{%
\includegraphics[width=70px]{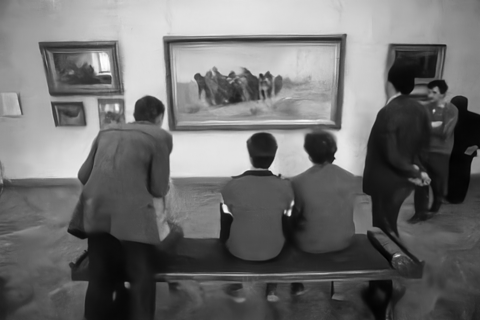}}~ \\[0.1cm]
\fbox{\includegraphics[width=70px]{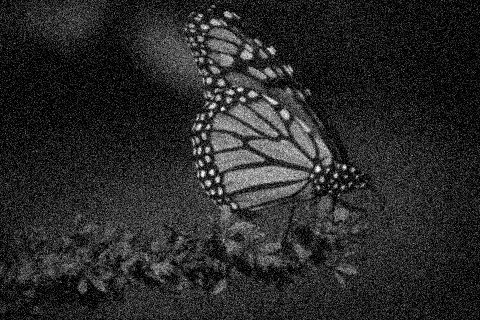}}~
\fbox{\includegraphics[width=70px]{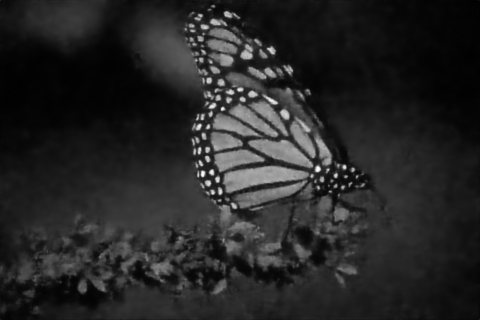}}~
\fbox{\includegraphics[width=70px]{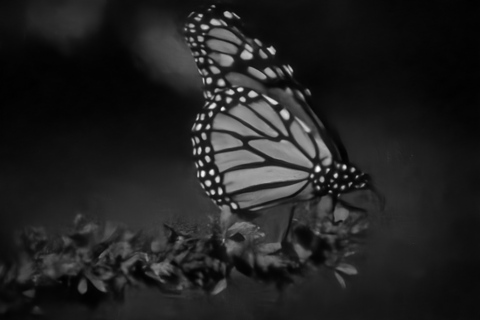}}~
\\[0.1cm]
\fbox{\includegraphics[width=70px]{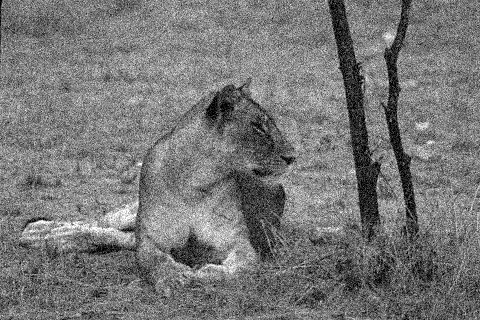}}~ \fbox{%
\includegraphics[width=70px]{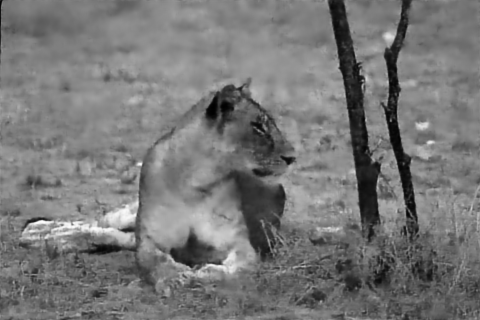}}~ \fbox{%
\includegraphics[width=70px]{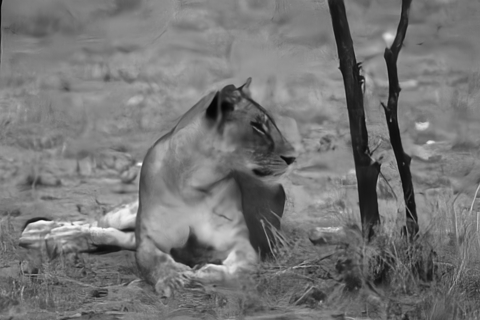}}~ \\[0.1cm]
\caption{Denoising results at $\sigma = 25$ on our noisy natural images simulated on the BSDS
dataset \protect\cite{bsds}. The input noisy image on the left, our results
in the middle and DnCNN \protect\cite{zhang2017beyond} on the right. As can
be seen, both methods achieve similar quality of denoising while ours
approach preserve more details of the image texture and edges.}
\label{fig:denoising}
\end{figure}

Figure \ref{fig:denoising} depicts the denoising results using the BSDS
dataset. It follows that we achieve similar denoising as the DnCNN \cite%
{zhang2017beyond}, while better preserving the textures details of the input
images. Table \ref{table:bsd68} shows the results of different approaches on
the BSD68 dataset, where it follows that we achieve state-of-the-art
perceptual SSIM score results.
\begin{table}[tbh]
\centering%
\begin{tabular}{|l|c|c|c|}
\hline
Algorithm & $\sigma = 15$ & $\sigma = 25$ & $\sigma = 50$ \\ \hline\hline
IDCNN-E & 31.00/\textbf{0.9} & 28.86/\textbf{0.85} & 25.95/\textbf{0.75} \\ \hline
IDCNN & 30.80/0.89 & 28.73/0.84 & 25.93/\textbf{0.75} \\ \hline
DnCNN & 31.74/\textbf{0.9} & 29.89/\textbf{0.85} & 25.69/0.71 \\ \hline
BM3D & 31.07/0.88 & 28.26/0.81 & 24.57/0.67 \\ \hline
%NLM & 26.99/0.69 \\ \hline
\end{tabular}%
\caption{Quantitative PSNR(dB) and SSIM results of denoising on the noisy natural images from
the dataset of \protect\cite{bsds}. Our approach,
using IDCNN for denoising achieves the high perceptual score of SSIM
\protect\cite{ssim}. In addition, our distortion score of PSNR is also
competitive relative to the state-of the art approach of DnCNN \protect\cite%
{zhang2017beyond}. Our edge preserving auxiliary loss improve the
performance of IDCNN in denoising in both measurements. The highest
SSIM scores are highlighted.}
\label{table:denoising}
\end{table}
\begin{table}[tbh]
\centering%
\begin{tabular}{|l|c|c|c|}
\hline
Algorithm & $\sigma = 15$ & $\sigma = 25$ & $\sigma = 50$\\ \hline\hline
IDCNN-E & 30.78/\textbf{0.9} & 28.61/\textbf{0.84} & 25.78/\textbf{0.74} \\ \hline
IDCNN & 30.51/0.89 & 28.53/\textbf{0.84} & 25.73/\textbf{0.74} \\ \hline
DnCNN & 31.73/\textbf{0.9} & 29.16/\textbf{0.84} & 26.23/0.71 \\ \hline
DeepAM \cite{deepam} & 31.68/0.89 & 29.21/0.82 & 26.24/0.72 \\ \hline
TRD \cite{trd} & 31.42/0.88 & 28.91/0.81 & 25.96/0.70 \\ \hline
MLP \cite{mlp} & - & 28.91/0.81 & 26.00/0.71 \\ \hline
CSF \cite{csf} & 31.24/0.87 & 28.91/0.81 & - \\ \hline
BM3D & 31.12/0.87 & 28.91/0.81 & 25.65/0.69 \\ \hline
\end{tabular}%
\caption{The average PSNR(dB) and SSIM results of different methods on the
BSD68 dataset. The highest SSIM scores are
highlighted.}
\label{table:bsd68}
\end{table}
\begin{figure}[tbh]
	\centering
	\fbox{\includegraphics[width=70px]{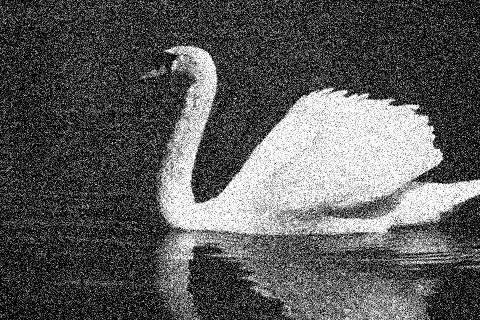}}~ \fbox{%
		\includegraphics[width=70px]{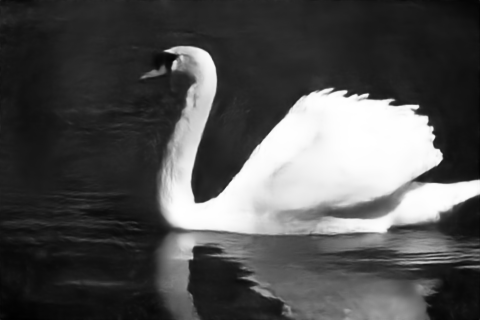}}~ \fbox{%
		\includegraphics[width=70px]{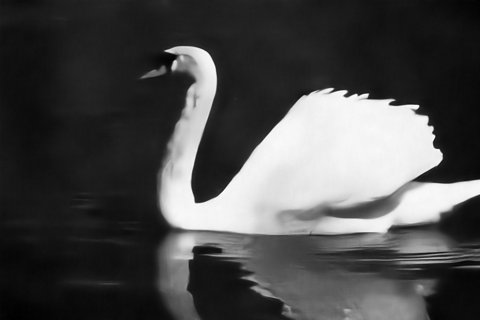}}~ \\[0.1cm]
	%\fbox{\includegraphics[width=70px]{Images/50_0001_1I.png}}~
	%\fbox{\includegraphics[width=70px]{Images/50_0001_1AUX.png}}~
	%\fbox{\includegraphics[width=70px]{Images/50_35049.png}}~
	%\\[0.1cm]
	\fbox{\includegraphics[width=70px]{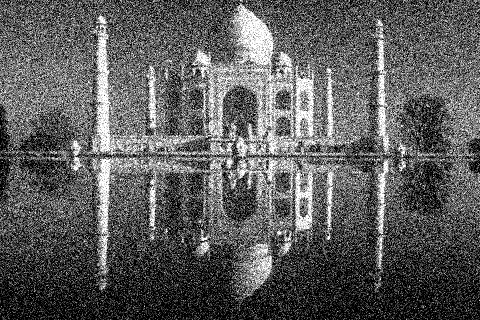}}~ \fbox{%
		\includegraphics[width=70px]{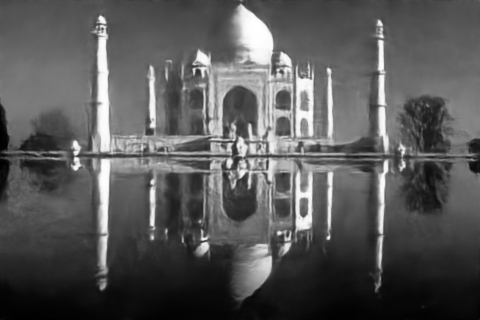}}~ \fbox{%
		\includegraphics[width=70px]{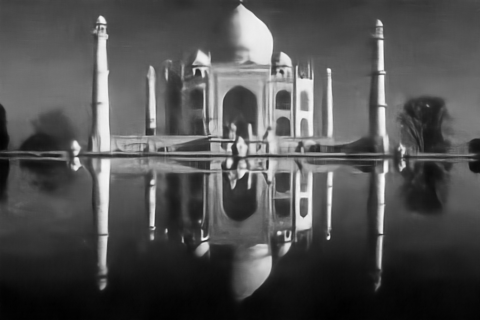}}~ \\[0.1cm]
	\caption{Denoising results at high $\sigma = 50$ on our noisy natural images simulated on the BSDS
		dataset \protect\cite{bsds}. The input noisy image on the left, our results
		in the middle and DnCNN \protect\cite{zhang2017beyond} on the right.}
	\label{fig:denoising50}
\end{figure}

\section{Conclusions}

\label{sec:conclusions}

We introduced a novel work for multi-scale processing of noisy images using edge preservation losses. Our
work is the first to solve detection of faint edges in noisy images by using deep learning
technique. We compared our method to FastEdges \cite{ofir2016fast} which is
the state-of-the-art in faint edge detection. We showed experimentally that
we succeed to improve this method in both aspects of run time and quality.
We achieved similar and better results on simulation and real images, while
improving the run times in orders of magnitude. FastEdges needed seconds to
process an image whereas ours algorithm requires only milli-seconds by
utilizing a fully convolutional network running on a GPU. Moreover, we
showed that our approach to overcome noise using deep multi-scale
preprocessing of the image, also improves the robustness of classifiers to
noisy objects. The accuracy of noisy objects classification increases
dramatically when applying the same preprocessing of our faint-edges
detector. We emphasize the robustness of the our CNN to noise also by the
classical problem of image denoising. Our approach to train this network for
denoising, using edge preservation auxiliary loss, achieves state-of-the-art
scores of noise removal in natural images.

{\small
\bibliographystyle{ieee}
\bibliography{egbib}

\begin{thebibliography}{10}\itemsep=-1pt

\bibitem{bsds}
P.~Arbelaez, M.~Maire, C.~Fowlkes, and J.~Malik.
\newblock Contour detection and hierarchical image segmentation.
\newblock {\em IEEE Trans. Pattern Anal. Mach. Intell.}, 33(5):898--916, May
  2011.

\bibitem{mcg}
P.~Arbel{\'a}ez, J.~Pont-Tuset, J.~T. Barron, F.~Marques, and J.~Malik.
\newblock Multiscale combinatorial grouping.
\newblock In {\em Proceedings of the IEEE conference on computer vision and
  pattern recognition}, pages 328--335, 2014.

\bibitem{blau2018perception}
Y.~Blau and T.~Michaeli.
\newblock The perception-distortion tradeoff.
\newblock In {\em Proceedings of the IEEE Conference on Computer Vision and
  Pattern Recognition}, pages 6228--6237, 2018.

\bibitem{nlm}
A.~Buades, B.~Coll, and J.-M. Morel.
\newblock A non-local algorithm for image denoising.
\newblock In {\em Computer Vision and Pattern Recognition, 2005. CVPR 2005.
  IEEE Computer Society Conference on}, volume~2, pages 60--65. IEEE, 2005.

\bibitem{mlp}
H.~C. Burger, C.~J. Schuler, and S.~Harmeling.
\newblock Image denoising: Can plain neural networks compete with bm3d?
\newblock In {\em 2012 IEEE conference on computer vision and pattern
  recognition}, pages 2392--2399. IEEE, 2012.

\bibitem{canny1987computational}
J.~Canny.
\newblock A computational approach to edge detection.
\newblock In {\em Readings in Computer Vision}, pages 184--203. Elsevier, 1987.

\bibitem{trd}
Y.~Chen, W.~Yu, and T.~Pock.
\newblock On learning optimized reaction diffusion processes for effective
  image restoration.
\newblock In {\em Proceedings of the IEEE conference on computer vision and
  pattern recognition}, pages 5261--5269, 2015.

\bibitem{bm3d}
K.~Dabov, A.~Foi, V.~Katkovnik, and K.~Egiazarian.
\newblock Image denoising with block-matching and 3d filtering.
\newblock In {\em Image Processing: Algorithms and Systems, Neural Networks,
  and Machine Learning}, volume 6064, page 606414. International Society for
  Optics and Photonics, 2006.

\bibitem{dollar2013structured}
P.~Doll{\'a}r and C.~L. Zitnick.
\newblock Structured forests for fast edge detection.
\newblock In {\em Computer Vision (ICCV), 2013 IEEE International Conference
  on}, pages 1841--1848. IEEE, 2013.

\bibitem{galun2007multiscale}
M.~Galun, R.~Basri, and A.~Brandt.
\newblock Multiscale edge detection and fiber enhancement using differences of
  oriented means.
\newblock In {\em Computer Vision, 2007. ICCV 2007. IEEE 11th International
  Conference on}, pages 1--8. IEEE, 2007.

\bibitem{sobel}
W.~Gao, X.~Zhang, L.~Yang, and H.~Liu.
\newblock An improved sobel edge detection.
\newblock In {\em Computer Science and Information Technology (ICCSIT), 2010
  3rd IEEE International Conference on}, volume~5, pages 67--71. IEEE, 2010.

\bibitem{logistic}
F.~E. Harrell.
\newblock Ordinal logistic regression.
\newblock In {\em Regression modeling strategies}, pages 311--325. Springer,
  2015.

\bibitem{horev2015detection}
I.~Horev, B.~Nadler, E.~Arias-Castro, M.~Galun, and R.~Basri.
\newblock Detection of long edges on a computational budget: A sublinear
  approach.
\newblock {\em SIAM Journal on Imaging Sciences}, 8(1):458--483, 2015.

\bibitem{crisp}
P.~Isola, D.~Zoran, D.~Krishnan, and E.~H. Adelson.
\newblock Crisp boundary detection using pointwise mutual information.
\newblock In {\em European Conference on Computer Vision}, pages 799--814.
  Springer, 2014.

\bibitem{deepam}
Y.~Kim, H.~Jung, D.~Min, and K.~Sohn.
\newblock Deeply aggregated alternating minimization for image restoration.
\newblock In {\em Proceedings of the IEEE Conference on Computer Vision and
  Pattern Recognition}, pages 6419--6427, 2017.

\bibitem{cifar}
A.~Krizhevsky, V.~Nair, and G.~Hinton.
\newblock The cifar-10 dataset.
\newblock {\em online: http://www. cs. toronto. edu/kriz/cifar. html}, 2014.

\bibitem{ssd}
W.~Liu, D.~Anguelov, D.~Erhan, C.~Szegedy, S.~Reed, C.-Y. Fu, and A.~C. Berg.
\newblock Ssd: Single shot multibox detector.
\newblock In {\em European conference on computer vision}, pages 21--37.
  Springer, 2016.

\bibitem{cob}
K.-K. Maninis, J.~Pont-Tuset, P.~Arbel{\'a}ez, and L.~Van~Gool.
\newblock Convolutional oriented boundaries: From image segmentation to
  high-level tasks.
\newblock {\em IEEE transactions on pattern analysis and machine intelligence},
  40(4):819--833, 2018.

\bibitem{marr1980theory}
D.~Marr and E.~Hildreth.
\newblock Theory of edge detection.
\newblock {\em Proc. R. Soc. Lond. B}, 207(1167):187--217, 1980.

\bibitem{wavelet_denoising}
M.~K. Mihcak, I.~Kozintsev, K.~Ramchandran, and P.~Moulin.
\newblock Low-complexity image denoising based on statistical modeling of
  wavelet coefficients.
\newblock {\em IEEE Signal Processing Letters}, 6(12):300--303, 1999.

\bibitem{ofir2016fast}
N.~Ofir, M.~Galun, B.~Nadler, and R.~Basri.
\newblock Fast detection of curved edges at low snr.
\newblock In {\em Proceedings of the IEEE Conference on Computer Vision and
  Pattern Recognition}, pages 213--221, 2016.

\bibitem{unet}
O.~Ronneberger, P.~Fischer, and T.~Brox.
\newblock U-net: Convolutional networks for biomedical image segmentation.
\newblock In {\em International Conference on Medical image computing and
  computer-assisted intervention}, pages 234--241. Springer, 2015.

\bibitem{schmidt2014shrinkage}
U.~Schmidt and S.~Roth.
\newblock Shrinkage fields for effective image restoration.
\newblock In {\em Proceedings of the IEEE Conference on Computer Vision and
  Pattern Recognition}, pages 2774--2781, 2014.

\bibitem{csf}
U.~Schmidt and S.~Roth.
\newblock Shrinkage fields for effective image restoration.
\newblock In {\em Proceedings of the IEEE Conference on Computer Vision and
  Pattern Recognition}, pages 2774--2781, 2014.

\bibitem{svm}
J.~A. Suykens and J.~Vandewalle.
\newblock Least squares support vector machine classifiers.
\newblock {\em Neural processing letters}, 9(3):293--300, 1999.

\bibitem{resnet}
C.~Szegedy, S.~Ioffe, V.~Vanhoucke, and A.~A. Alemi.
\newblock Inception-v4, inception-resnet and the impact of residual connections
  on learning.
\newblock In {\em AAAI}, volume~4, page~12, 2017.

\bibitem{unnikrishnan2005measure}
R.~Unnikrishnan, C.~Pantofaru, and M.~Hebert.
\newblock A measure for objective evaluation of image segmentation algorithms.
\newblock In {\em Computer Vision and Pattern Recognition-Workshops, 2005. CVPR
  Workshops. IEEE Computer Society Conference on}, pages 34--34. IEEE, 2005.

\bibitem{wang2017detecting}
Y.-Q. Wang, A.~Trouv{\'e}, Y.~Amit, and B.~Nadler.
\newblock Detecting curved edges in noisy images in sublinear time.
\newblock {\em Journal of Mathematical Imaging and Vision}, 59(3):373--393,
  2017.

\bibitem{ssim}
Z.~Wang, A.~C. Bovik, H.~R. Sheikh, and E.~P. Simoncelli.
\newblock Image quality assessment: from error visibility to structural
  similarity.
\newblock {\em IEEE transactions on image processing}, 13(4):600--612, 2004.

\bibitem{hed}
S.~Xie and Z.~Tu.
\newblock Holistically-nested edge detection.
\newblock In {\em Proceedings of the IEEE international conference on computer
  vision}, pages 1395--1403, 2015.

\bibitem{cedn}
J.~Yang, B.~Price, S.~Cohen, H.~Lee, and M.-H. Yang.
\newblock Object contour detection with a fully convolutional encoder-decoder
  network.
\newblock In {\em Proceedings of the IEEE Conference on Computer Vision and
  Pattern Recognition}, pages 193--202, 2016.

\bibitem{zhang2017beyond}
K.~Zhang, W.~Zuo, Y.~Chen, D.~Meng, and L.~Zhang.
\newblock Beyond a gaussian denoiser: Residual learning of deep cnn for image
  denoising.
\newblock {\em IEEE Transactions on Image Processing}, 26(7):3142--3155, 2017.

\end{thebibliography}
}

\end{document}